\documentclass[conference]{IEEEtran}
\IEEEoverridecommandlockouts
\usepackage{cite}
\usepackage{amsmath,amssymb,amsfonts}
\usepackage{algorithmic}
\usepackage{graphicx}
\usepackage{hyperref}
\usepackage{textcomp}
\usepackage{xcolor}
\usepackage{xspace}
\usepackage{lipsum}
\usepackage{subcaption} 
\usepackage{array} 
\usepackage{multirow}

\newcommand{\methodname}{TrajGenAgent\xspace}

\usepackage{balance}
\usepackage{todonotes}
\usepackage{newfloat}
\usepackage{listings}
\usepackage{comment}
\usepackage{url}
\usepackage{makecell}
\usepackage{booktabs}

\def\BibTeX{{\rm B\kern-.05em{\sc i\kern-.025em b}\kern-.08em
    T\kern-.1667em\lower.7ex\hbox{E}\kern-.125emX}}
\begin{document}

\title{TrajGenAgent: A 
Hierarchical LLM Agent 
for 
Human Mobility Trajectory Generation\\

}

\author{
\IEEEauthorblockN{
Siyu Li\IEEEauthorrefmark{2},
Toan Tran\IEEEauthorrefmark{2},
Lingyi Zhao\IEEEauthorrefmark{3},
Khurram Shafique\IEEEauthorrefmark{3},
Li Xiong\IEEEauthorrefmark{2}
}

\IEEEauthorblockA{
\IEEEauthorrefmark{2}
Dept. of Computer Science, Emory University, Atlanta, GA, USA
}

\IEEEauthorblockA{
\IEEEauthorrefmark{3}
Novateur Research Solutions, Ashburn, VA, USA
}

\IEEEauthorblockA{
\{siyu.li, viet.toan.tran, lxiong\}@emory.edu, 
\{lzhao, kshafique\}@novateur.ai
}
}

\maketitle

\begin{abstract}

Human mobility data is essential for transportation, urban planning, and epidemic control. Yet large-scale trajectory collection is often costly and restricted by privacy concerns, motivating the need for realistic synthetic mobility trajectory generation. 
Recent LLM-based generators typically follow two paradigms: (i) prompt engineering, which provides efficient zero-shot generation with general prior knowledge but lacks fine-grained spatiotemporal grounding; and (ii) fine-tuning with structured trajectories, which achieves strong spatiotemporal precision but incurs substantial computational cost and may weaken general reasoning. Tool-augmented agents are emerging but remain at an early stage and still lack effective coordination between high-level planning and low-level realization. To address these limitations, we propose \methodname, a semantic-aware hierarchical LLM-agent framework for trajectory generation without model fine-tuning. \methodname\ adopts a two-stage orchestrator--worker design that decouples macro-level activity structure from micro-level spatiotemporal dynamics. In the first stage, an LLM synthesizes an activity chain for a given individual and day of the week via in-context learning over historic examples. 
In the second stage, a deterministic workflow instantiates each activity visit with distance-aware rule-based location retrieval and LLM augmented kinematics-aware temporal generation.

Traditional evaluation metrics for synthetic mobility data primarily assess aggregate spatiotemporal statistics, which do not capture behavioral fidelity or realism of individual trajectories. To address this limitation, we introduce an anomaly-detection-based evaluation framework with two complementary anomaly detectors that provide behavior \& semantic feedback beyond macro-level statistical consistency. Experiments on both benchmark and large-scale simulation datasets show that \methodname\ outperforms baselines in both spatiotemporal statistical metrics while also improving semantic coherence and individual-specific behavior fidelity, all without parameter updates.

\end{abstract}

\begin{IEEEkeywords}
Human Mobility Trajectory Generation, LLM Agent, Orchestrator--Worker Architecture, Zero-Shot Reasoning, Workflow-Based Tool Integration
\end{IEEEkeywords}

\begin{figure*}[ht]
    \centering
    \includegraphics[width=0.95\linewidth]{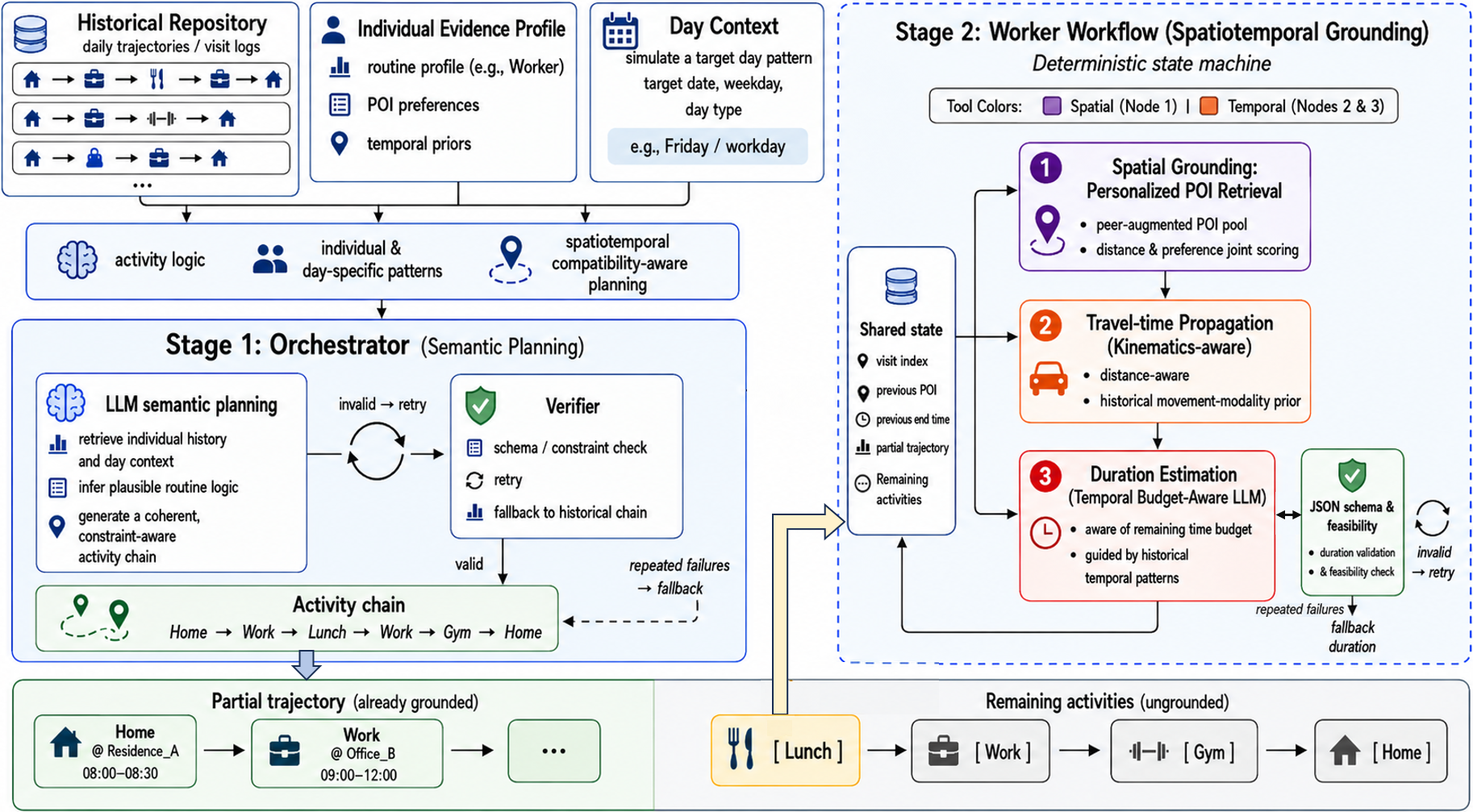}
    \vspace{-2mm}    
    \caption{The \methodname framework. A hierarchical orchestrator--worker LLM agent workflow, where a LLM orchestrator synthesizes individual- and day-controlled activity-chain scaffolds via in-context learning over historical evidence, and a deterministic LangGraph worker loop grounds each activity into complete visits by peer-augmented POI retrieval, distance-/kinematics-aware travel-time propagation, and a context-aware, constraint-guided LLM duration module.}
    \label{fig:trajgenagent}
    \vspace{-5.5mm}
\end{figure*}

\section{Introduction}

Human mobility data, represented as trajectories or sequences of visits, is essential for advancing research and applications in transportation, urban planning, social dynamics, and epidemiology \cite{hu2021human, becker2011tale, chen2019human}. However, collecting large-scale trajectory data is often constrained by high costs and privacy concerns, making real-world mobility datasets difficult to access \cite{10.1145/3652158}. This has motivated the development of synthetic yet realistic trajectory generation methods.

\smallskip
\noindent
\textbf{Existing Approaches and Limitations.}
Early approaches relied on micro-simulators calibrated using sensor data, traffic statistics, and handcrafted behavioral rules \cite{LE201681}. Such simulation-based methods require careful parameter tuning and often fail to capture complex mobility patterns due to over-simplified heuristics \cite{feng2020learning}. More recent data-driven models, including Generative Adversarial Networks (GAN) and spatiotemporal point processes, learn mobility distributions directly from data to generate large-scale trajectories \cite{lin2023generating,zhang2023csgan,ouyang2018non,long2023practical}. However, they often suffer from training instability, scalability limitations, remain limited in modeling long-range and complex dependencies, and lack explicit semantic understanding of human routines.

With the emergence of Large Language Models (LLMs) \cite{zhao2023survey}, recent studies have begun to leverage their powerful sequence modeling and reasoning capabilities for trajectory generation \cite{li2025geo, bhandari2024urban, zhang2025study}. Existing LLM-based methods can be categorized by how spatiotemporal knowledge is integrated: (1) language-level approaches, including pure prompting that relies on semantic priors \cite{jiawei2024large} and lightweight text-based fine-tuning that inject global spatiotemporal knowledge in natural language form \cite{bhandari2024urban}; (2) trajectory-level fine-tuning that encodes structured trajectories as token sequences \cite{li2025geo}; and (3) tool-augmented agent paradigms that externalizes spatiotemporal grounding by invoking dedicated tools or knowledge sources during inference time \cite{li2025towards, zhang2025study}.

Language-level approaches preserve an LLM’s general knowledge and adaptability, but offer limited capability of fine-grained spatiotemporal pattern injection, often yielding trajectories that are semantically plausible yet poorly calibrated in time and space. In contrast, trajectory-level fine-tuning methods such as Geo-Llama \cite{li2025geo} represent trajectories as sequences of visits and adapt pretrained LLMs via parameter-efficient fine-tuning (e.g., LoRA \cite{hu2022lora}). While achieving strong spatiotemporal fidelity, they introduce substantial computational overhead and tightly entangle semantic reasoning with low-level statistical pattern fitting, over-specializing the model and reducing general reasoning capacity and control of individual-level semantic behaviors. 


These observations reveal a fundamental tension: realistic mobility generation requires both semantic coherence and fine-grained spatiotemporal grounding. Here, semantic coherence refers to the consistency of a generated trajectory with human activity logic, individual- and day-specific routines, and activity--POI--time compatibility, while fine-grained grounding concerns the accurate reflection of location preferences, transition patterns, and temporal regularities. The above two approaches struggle to achieve both simultaneously. Tool-augmented agent paradigms, while promising, remain at an early stage and still lack effective coordination between high-level behavioral planning and low-level spatiotemporal realization.

\smallskip
\noindent
\textbf{Challenges in LLM Agent based Mobility Generation.}
LLM agent based paradigm offers a principled solution to the previously observed tension: the LLM serves as a semantic reasoning and planning core, while external tools can inject precise spatiotemporal evidence at inference time without requiring all domain knowledge to be encoded into model parameters \cite{yao2022react, schick2023toolformer}. However, effectively leveraging LLM agents for trajectory generation presents several challenges.

General LLM agent frameworks primarily differ in how they  structure and enforce tool invocation: 1) schema-registered prompt-based calling, 2) supervised fine-tuning on function-call traces for higher reliability (e.g., tool-calling specialized models) \cite{zhang2024xlam}, and 3) workflow-managed agents that organize tool execution as a state machine to guarantee stable control flow and termination while reserving the LLM for steps requiring semantic generalization \cite{wang2024langgraph}.

A natural agent design for trajectory generation is to fine-tune or prompt an LLM to autonomously invoke tools via structured function-call schemas \cite{zhang2024xlam}.
However, trajectory generation requires repeated, deterministic decisions for every visit (location choice, travel-time propagation, duration estimation), which makes end-to-end autonomous tool calling fragile. Schema compliance cannot be guaranteed at 100\% for long-horizon trajectories; a single malformed or missed tool call can leave missing fields (e.g., POI or time) and break visit-to-visit dependencies, trigger cascading errors  that corrupt subsequent steps and the overall daily schedule.
Moreover, supervised fine-tuning for tool-calling behavior introduces additional training cost and may compromise generalization capability due to catastrophic forgetting \cite{li2025towards}.

Workflow-managed orchestration offers a promising alternative for stable and deterministic tool execution in long-horizon trajectory generation. However, designing such workflows is nontrivial: it requires structuring control flow, enforcing visit-level dependencies, and balancing deterministic execution with semantic flexibility. To our knowledge, prior work has not explored workflow design for trajectory generation.

\smallskip
\noindent
\textbf{Evaluation Gap in Mobility Generation.}
Beyond agent design, a complementary challenge lies in how generated trajectories are evaluated.  
Most trajectory-generation evaluations rely on aggregate distributional distance metrics (e.g., Jensen–Shannon divergence (JSD) between real training data and generated trajectories over travel distance, visit frequency, or transition matrices) \cite{zhang2023csgan, li2025geo}. These metrics quantify population-level statistical similarity but often miss individual-level semantic defects---e.g., a location--time pattern that is normal for one may be anomalous for another, even if global statistics match.

\smallskip
\noindent
\textbf{Contributions.}
We propose \methodname, a zero-shot hierarchical agent framework that orchestrates heterogeneous reasoning components within a deterministic workflow implemented using LangGraph\cite{wang2024langgraph}. By integrating LLM-based reasoning with rule-based retrieval and explicit physical calculations, it ensures semantic-aware planning and physics-aware fine-grained spatiotemporal knowledge injection without costly model updates. As shown in Fig. \ref{fig:trajgenagent}, \methodname\ decomposes generation into a macro-to-micro pipeline. In Stage 1, an orchestrator LLM  produces an activity-chain scaffold (a semantic skeleton) via in-context learning over an individual's historical daily chains and contextual information (personal attributes and day context). In Stage 2, specialized worker modules transform each activity into a complete visit  through a predefined workflow that injects fine-grained spatiotemporal knowledge. 

Concretely, Stage 2 consists of two specialized workers: a spatial worker and a temporal worker.  The spatial worker performs rule-based location retrieval using personal statistical priors from a peer-augmented candidate pool constructed by similarity matching across individuals. This enables controlled exploration while restricting locations within a personalized feasible set. A distance-aware mechanism further enforces transition plausibility by aligning candidate locations with the user’s historical activity-pair moving modalities, ensuring consistency with observed velocity and movement distributions. 
The temporal worker advances time by jointly inferring arrival timestamps and stay durations. Specifically, it combines (i) a kinematics-aware travel-time estimator that leverages moving-modality priors with (ii) an LLM-based duration module that respects time budget constraints. Given the generated history and the remaining itinerary, the workflow iteratively calibrates each visit’s arrival time and dwell time to maintain local transition plausibility and day-level schedule consistency. To ensure robustness, lightweight verifiers supervise both workers through schema-enforced fallbacks and feasibility constraints, which ensure both structural format and time validity.


Finally, to better evaluate behavior-level plausibility beyond traditional aggregate statistical metrics, we propose a novel anomaly-detection based evaluation framework. We use two detectors with complementary emphases: ICAD which identifies local visit-wise inconsistencies \cite{azarijoo2025icad} and BeSTAD which captures user-level behavioral shifts \cite{xie2025bestad}. Both are applied post hoc to assess the semantic coherence of generated trajectories. 

Our contributions are summarized as follows:
\begin{itemize}
    \item \textbf{Hierarchical LLM-Agent Framework}. We propose \methodname, a zero-shot hierarchical LLM-agent framework that injects spatiotemporal knowledge at inference time via a deterministic, verifier-guarded orchestrator--worker workflow. By separating macro-level activity-chain planning from visit-level grounding, \methodname\ enables high-fidelity trajectory generation without costly fine-tuning or fragile autonomous tool calling.
    
    \item \textbf{Personalized and Physics-Aware Control}. We enable fine-grained personalized control over user- and day-specific routines by constraining generation with historical evidence and configurable tool scopes, while enforcing time budgets and physics-aware mobility during spatiotemporal grounding. This yields trajectories that are both semantically coherent and faithful to fine-grained spatiotemporal statistics.

    \item \textbf{Behavior-Aware Evaluation Framework}. We introduce a novel evaluation framework that augments traditional statistical metrics with two complementary anomaly detectors (ICAD and BeSTAD), trained to distinguish real from abnormal or implausible trajectories, to assess the semantic coherence of the generated trajectories.

    \item \textbf{Comprehensive Experimental Evaluation.} Experiments on large scale datasets show that \methodname\ outperforms baselines on spatiotemporal statistical alignment and semantic coherence, without expensive parameter updates. Anomaly-detection results suggest that our inference-time grounding with individualized evidence and kinematics-aware priors preserves semantic plausibility and avoids detectable artifacts, while baselines can still exhibit anomalous trajectory patterns despite matching aggregate statistics, especially on behaviorally diverse datasets.
\end{itemize}

\section{Related Work}

\noindent
\textbf{Mobility Trajectory Generation without LLMs.}
Prior to LLM-based approaches, mobility generation was dominated by simulation and neural-based generative modeling.
Simulation-based methods synthesize trajectories using hand-crafted behavioral rules and physically motivated estimations calibrated from sensors or surveys \cite{pelekis2013hermoupolis,jiang2016timegeo}. They are often brittle due to over-simplified heuristics.
Data-driven models instead learn trajectory distributions directly from historical data.
A common formulation encodes mobility as a \emph{fixed-interval} spatiotemporal sequence, and learns next-step transitions with recurrent backbones (e.g., RNN/LSTM variants) \cite{liu2016predicting,feng2018deepmove}.
Beyond one-step predictors, GAN-based generators improve distribution matching by adversarial training \cite{goodfellow2020generative}, with representative variants including adversarial trajectory synthesis \cite{ouyang2018non} and reinforcement-learning-based sequence generation \cite{yu2017seqgan,feng2020learning,zhang2023csgan}.
Despite progress, fixed-interval representations produce unnecessarily long sequences with repeated location states and implicit time encoding, resulting in ambiguous temporal representation and degrades trajectory generation quality. To address these limitations, recent work adopts \emph{visit-wise} trajectory and continuous-spatiotemporal formulations based on deep spatiotemporal point processes (DeepSTPP) \cite{zhou2022neural,long2023practical}, which model trajectories as irregular visit sequences and jointly capture where and when visits occur. However, these generators remain limited in modeling  long-range, complex dependencies and lack explicit semantic understanding of human routines and behavioral logic. 

\vspace{0.7mm}
\noindent
\textbf{LLM-based Mobility Trajectory Generation.}
Recent studies leverage LLMs for mobility generation by exploiting their strong sequence modeling and semantic reasoning capabilities. They mainly differ in \emph{how} fine-grained spatiotemporal knowledge is incorporated.

\paragraph{Pure prompting-based generation}
A representative line of work relies on pure prompt engineering to generate plausible trajectories in a zero-shot manner \cite{jiawei2024large}. For example, \cite{bhandari2024urban} crafts prompts using statistical summaries (e.g., demographics, event types, and event--temporal correlations) to emulate travel-diary-style generation without additional training. \cite{wang2024large} further improves prompting via self-consistent activity pattern identification and retrieval-augmented generation (RAG), using LLM priors to evaluate candidate prompt combinations conditioned on individual profiles and POI background information.
While prompting can exploit rich contextual cues to produce semantically coherent, narrative-level mobility routines, it typically lacks explicit injections of fine-grained spatiotemporal information, which limits precise micro-level spatiotemporal grounding.

\paragraph{Trajectory-level fine-tuning}
In contrast, trajectory-level fine-tuning injects spatiotemporal knowledge into model parameters by encoding structured trajectories as visit sequence prompts and fine-tuning pretrained LLMs with parameter-efficient adapters such as LoRA \cite{hu2022lora}. Geo-Llama \cite{li2025geo} exemplifies this paradigm by optimizing a next-token prediction objective over discretized spatiotemporal visit tokens, and further introducing visit-wise permutation so the model learns temporal regularities from time features within visits rather than from the original sequence order, yielding strong micro-level spatiotemporal fidelity without external semantic annotations. However, heavy adaptation on structured tokens can introduce substantial computational overhead and entangle semantic reasoning with low-level spatiotemporal statistics,  reducing the general semantic capabilities and flexibility of foundation models.

\paragraph{LLM agents and tool-augmented workflows}
The above limitations motivate an emerging paradigm that reframes LLMs as \emph{tool-augmented agents} rather than monolithic generators, enabling inference-time spatiotemporal knowledge injection through modular interfaces \emph{without costly fine-tuning or parameter updates} \cite{yao2022react,schick2023toolformer,zhang2025study}. Beyond domain knowledge, robust agentic generation often requires \emph{procedural} control over structured reasoning (e.g., CoT-style traces \cite{wei2022chain}) and tool usage (e.g., schema-registered function calls \cite{schick2023toolformer}).
Existing agent frameworks typically enforce tool use via three strategies: (i) supervised fine-tuning on tool-call traces with explicit JSON/function-call schemas to improve invocation reliability (e.g., tool-calling specialized models such as xLAM \cite{zhang2024xlam}); (ii) zero-shot prompting with schema/tool registration, leveraging strong foundation models that can follow JSON-style interfaces (e.g., GPT4-OSS-120B \cite{gpt4oss2024}); and (iii) workflow-managed orchestration that encodes tool execution as an explicit state machine to guarantee stable control flow and termination (e.g., LangGraph \cite{wang2024langgraph}).
Such workflow-managed agents can flexibly compose heterogeneous tools 
to inject spatiotemporal evidence at inference time while preserving the LLM as a semantic planner.
However, systematically designing reliable, long-horizon workflows for mobility generation remains relatively under-explored \cite{zhang2025study}. 

\vspace{0.7mm}
\noindent
\textbf{GPS Trajectory Generation and POI Recommendation.}
Apart from human mobility trajectory generation, two closely related topics study synthetic movement data under different objectives and representations.

\emph{GPS trajectory generation} targets dense coordinate streams at very fine-grained spatial and temporal resolution (e.g., per-second), instead of visit-wise trajectories that encode discrete activities and human behavioral semantics. Early works synthesize GPS traces by perturbing real trajectories or recombining trajectory segments \cite{armstrong1999geographically,zandbergen2014ensuring}, which can distort spatiotemporal characteristics and reduce utility.
Data-driven approaches have also been explored, including GAN-based approaches \cite{wang2021large,cao2021generating} and diffusion-based models such as DiffTraj \cite{zhu2023difftraj}, which learn fine-grained spatiotemporal dynamics from raw GPS sequences.
Due to this distinct granularity and goal, we do not include GPS-trajectory generators in our comparisons.

\emph{Next point-of-interest (POI) recommendation} predicts an individual's next POI conditioned on historical mobility and context, typically framed as a sequential recommendation problem, with limited or coarse temporal modeling. Prior work spans probabilistic, deep learning, graph-based, and LLM-based recommenders \cite{cheng2013you,kong2018hst,sun2020go,luo2021stan,zhang2022next,lim2020stp,yang2022getnext,li2024large}. In contrast, our task is \emph{trajectory generation}: synthesizing full-day visit sequences with both visit-specific spatiotemporal fidelity and realistic global dynamics, rather than focusing on one-step recommendation accuracy.
Although recommenders can be rolled out autoregressively, they often lack explicit mechanisms to maintain long-horizon coherence.
Given these fundamental differences, we do not treat POI recommendation methods as baselines for our work.

\section{TrajGenAgent Framework}
\label{sec:method}

\subsection{Trajectory Representation and Problem Setup}
We represent a daily trajectory as a sequence of visits rather than a fixed-interval time series.  
For an individual $u$ on date $d$, a trajectory is a sequence of visits:
\begin{equation}
\mathcal{T}_{u,d}=\left[(a_i, p_i, t_i^{s}, t_i^{e})\right]_{i=1}^{N_{u,d}},
\end{equation}
where $a_i\in\mathcal{A}$ is the activity type, $p_i\in\mathcal{P}$ is a POI identifier (with associated latitude/longitude), and $t_i^{s}, t_i^{e}$ are the start/end timestamps of the visit.
The number of visits $N_{u,d}$ varies by individual and day, and we further define the visit duration as:
\begin{equation}
\delta_i = t_i^{e}-t_i^{s}.
\end{equation}

Given a historical trajectory dataset, an individual $u$ and target date $d$, our goal is to generate a visit sequence $\mathcal{T}_{u,d}$ that is realistic, reflecting both the individual's historical mobility and population-level behavioral patterns.

\subsection{TrajGenAgent Overview}
\methodname\ adopts a hierarchical \textit{orchestrator--worker} agent architecture implemented as a deterministic workflow in LangGraph \cite{wang2024langgraph}. 
Given an individual $u$ and target date $d$, generation is decomposed into two stages:
\begin{equation}
(u,d,\mathcal{H}_u) \xrightarrow{\text{Stage 1: Orchestrator}} \mathcal{C}_{u,d}
\xrightarrow{\text{Stage 2: Worker Workflow}} \mathcal{T}_{u,d},
\end{equation}
where $\mathcal{H}_u$ is the historical repository, $\mathcal{C}_{u,d}=[a_1,\ldots,N_{u,d}]$ is an activity-chain \emph{semantic skeleton}, and $\mathcal{T}_{u,d}$ is the final visit trajectory.

\textbf{Stage 1 (Orchestrator)} performs \emph{semantic planning}: it prompts an LLM with individual-conditioned historical evidence (exemplar daily chains and compact statistical summaries) and synthesizes a plausible activity chain under hard lexical and structural constraints.
\textbf{Stage 2 (Worker Workflow)} performs \emph{spatiotemporal grounding}: it deterministically instantiates each activity into $(p_i,t_i^s,t_i^e)$ by executing a fixed sequence of modules (POI retrieval, travel-time propagation, and duration estimation) until completion. In our implementation, both stages share the same instruction-tuned backbone Qwen2.5-32B-Instruct \cite{yang2025qwen3}, served via a vLLM \cite{kwon2023efficient} inference server exposing an OpenAI API--compatible interface for high-throughput generation in the deterministic workflow.

\paragraph{Why deterministic workflow instead of free-form tool calling}
A natural alternative is to fine-tune an LLM to autonomously invoke MCP-style tools through schema-constrained outputs.
We avoid this design for three reasons: (i) \emph{reliability}---schema fine-tuning still cannot guarantee 100\% valid calls, while mobility generation repeatedly makes structured decisions at every visit; (ii) \emph{efficiency}---tool-calling fine-tuning is compute-intensive and may weaken instruction-following robustness, whereas \methodname\ leverages strong off-the-shelf instruction models at inference time; and (iii) \emph{control/termination}---free-form autonomous calling can drift from global constraints, repeat locally, or loop, which is especially harmful for long-horizon daily sequences. Encoding the procedure as an explicit state machine yields bounded execution, predictable control flow, and reproducibility.

\paragraph{Activity chain as a stabilizing intermediate}
Introducing $\mathcal{C}_{u,d}$ decouples \emph{semantic planning} from \emph{spatiotemporal realization}, which simplifies long-horizon generation and reduces error accumulation across visits.
As a high-level scaffold, $\mathcal{C}_{u,d}$ anchors global day structure while allowing the worker to inject deterministic mobility priors (e.g., distance- and speed-based feasibility) without requiring the LLM to directly rank or search over large POI candidate sets.

\paragraph{Evidence-driven decisions with lightweight verifiers}
Across both stages, \methodname follows an evidence-to-decision pattern: historical observations provide \emph{evidence} (activity and transition tendencies, POI preferences, duration statistics, and mobility priors), and workflow modules translate this evidence into constrained \emph{decisions}.
Lightweight \emph{verifiers} enforce strict output schemas and feasibility bounds through bounded checks and clipping; upon violations, the workflow triggers deterministic repair or fallback to preserve structural validity and schedule feasibility.

Overall, \methodname balances foundation-model semantic generalization with deterministic spatiotemporal grounding, enabling reliable large-scale trajectory generation without fine-tuning while remaining extensible to richer semantic controls and evaluator-in-the-loop feedback.

\subsection{Stage 1: Orchestrator for Activity-Chain Generation}
\label{subsec:stage1}

\begin{figure}[t]
    \centering
    \vspace{-2mm}
    \includegraphics[width=\columnwidth]{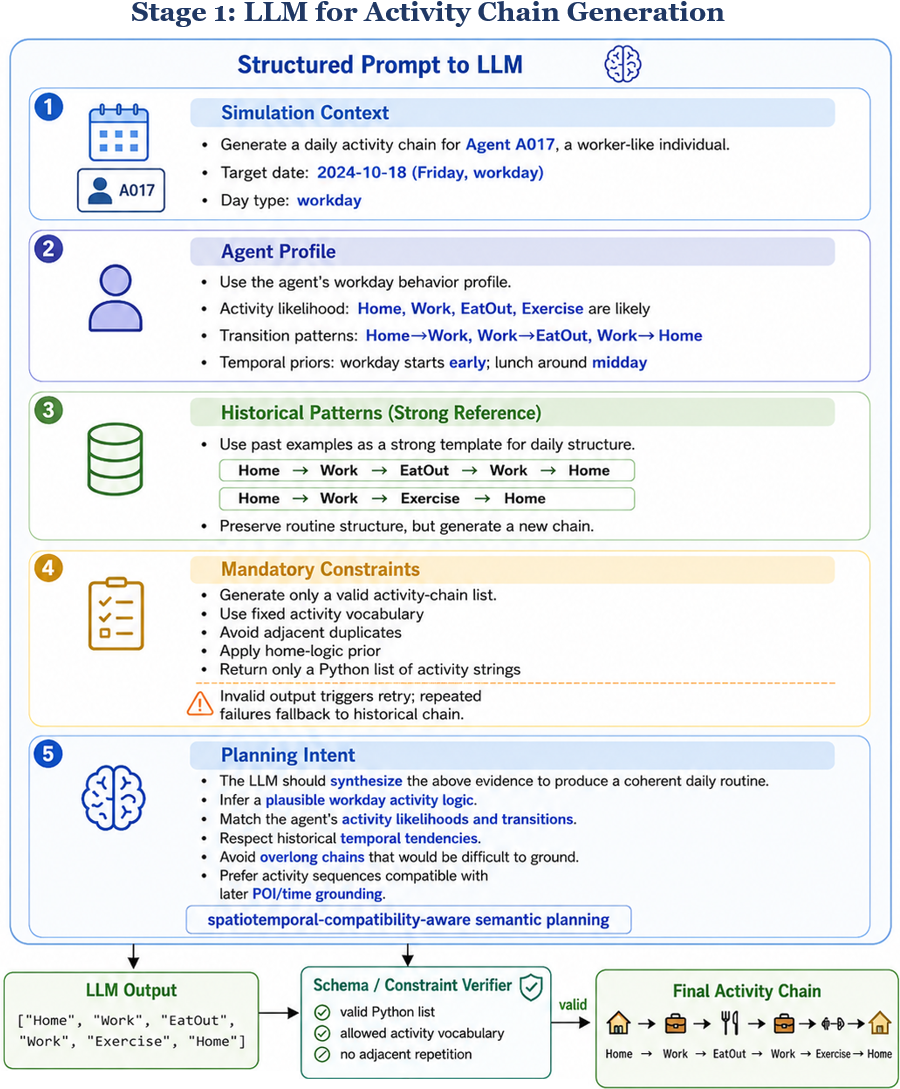}
    \caption{The illustration of Stage~1 LLM semantic planning for activity-chain generation.}
    \label{fig:semantic_planning_prompt}
    \vspace{-6mm}
\end{figure}

Stage~1 generates a daily activity chain $\mathcal{C}_{u,d}=[a_1,\ldots,a_{N_{u,d}}]$ as a \emph{semantic plan} for individual $u$ on date $d$.
Rather than training a dedicated generator, the orchestrator leverages \emph{in-context learning} (ICL) over \emph{individual-conditioned evidence} at inference time, enabling fine-grained personalization even in narrow regimes (e.g., conditioning on a specific weekday or day type). Figure~\ref{fig:semantic_planning_prompt} illustrates the prompt-level semantic planning process used by the orchestrator.

From each individual's historical trajectories, we construct an evidence profile $\Pi_u$ that characterizes individual preferences and mobility regularities.
In addition to activity- and POI-level statistics, $\Pi_u$ includes transition-level mobility priors derived from consecutive visits using the great-circle distance $\ell_{i-1,i}=\mathrm{dist}(p_{i-1},p_i)$ and the observed inter-visit gap $t_i^{s}-t_{i-1}^{e}$.

\vspace{-4mm}

\begin{equation}
\begin{aligned}
\Pi_u = \Big\{&
\pi_u(a),\;
\pi_u(a'\!\mid a),\;
\pi_u(p\!\mid a),\;
\mu_u(\delta\!\mid a),\\
&
\mu_u\!\big(\ell\mid a\!\rightarrow\!a'\big),\;
\mu_u\!\big(v\mid a\!\rightarrow\!a'\big),\;
\mu_u(t_1^{s}\!\mid w,a)
\Big\},
\end{aligned}
\end{equation}
where $\pi_u(a)$ is the activity occurrence likelihood, $\pi_u(a'\!\mid a)$ is the activity transition tendency, and $\pi_u(p\!\mid a)$ is the activity-conditioned POI preference. Here, $\pi_u(\cdot)$ denotes empirical categorical distributions over discrete choices, while $\mu_u(\cdot)$ denotes empirical priors over continuous-valued quantities. The remaining terms characterize empirical priors of visit duration, transition distance, transition speed, and weekday-conditioned first-start-time patterns (with $w$ denoting the day type or day-of-week).

\subsubsection{Evidence construction from individual history}
We construct daily activity chains from the individual-specific historical repository $\mathcal{H}_u$ by extracting each day’s visit sequence in temporal order.
For a target date $d$, we retrieve a date-specific evidence subset $\mathcal{E}_{u,d}\subset\mathcal{H}_u$ using a prioritized policy:
(i) same weekday, (ii) same day type (weekday/weekend), and (iii) fallback to all available days.
From $\mathcal{E}_{u,d}$ we construct two complementary evidence views:

\begin{itemize}
    \item \textbf{Exemplar evidence (ICL anchors):} a small set of historical chains presented verbatim as strong references, allowing the LLM to induce day structure via in-context pattern learning.
    \item \textbf{Summary evidence (compact priors):} compact activity-frequency and transition-tendency summaries derived from the evidence profile $\Pi_u$ in Eq.~(4), which regularize generation and reduce implausible chains.
\end{itemize}

\subsubsection{Evidence-based activity chain generation with verification}
The orchestrator LLM is prompted with (1) \emph{day-level controllable signals} (e.g., day-of-week and day type), (2) \emph{historical evidence} in the form of exemplar activity chains and compact statistical priors (e.g., activity frequencies and transition tendencies), and (3) hard constraints, including a fixed activity vocabulary and a no-adjacent-duplicate rule (with an optional soft home-start/end prior). The output is restricted to a Python list of activity strings. To ensure robustness, we apply a bounded \emph{generate--verify} loop:
\begin{itemize}
    \item \textbf{Schema check:} parse the output as a list and reject malformed generations.
    \item \textbf{Constraint check:} enforce vocabulary membership, and no-adjacent-duplicate constraints.
    \item \textbf{Repair/fallback:} retry briefly; if still invalid, fall back to an evidence-derived default (e.g., a sampled historical chain) to prevent error propagation to Stage~2.
\end{itemize}

\subsubsection{Why evidence-based activity chain generation outperforms fine-tuning for personalization}
Compared to fine-tuned generators, this orchestrator is \textbf{data-adaptive}---it can specialize to a single individual or a weekday-specific routine simply by adjusting the evidence scope---and offers \textbf{fine-grained controllability} through the evidence selection policy. Moreover, by leveraging prompting-based in-context learning, it preserves the backbone model's semantic generalization and requires only inference-time compute (no parameter updates), yielding a personalized and semantically coherent scaffold for Stage~2 grounding.

\vspace{-2mm}

\subsection{Stage 2: Workflow-based Spatiotemporal Grounding}
\label{subsec:stage2}

Given the semantic skeleton $\mathcal{C}_{u,d}=[a_1,\ldots,a_{N_{u,d}}]$, \methodname\ finalizes a complete visit trajectory by executing a deterministic state-machine workflow (implemented in LangGraph)  over $i=1\ldots N_{u,d}$. The worker workflow maintains a shared state (e.g., visit index, previous POI, current time, and partial trajectory) that each module reads and updates to propagate visit-to-visit dependencies, and then applies a fixed module order for each visit:
\begin{equation}
\begin{aligned}
\texttt{location\_node} &\rightarrow \texttt{travel\_time\_node} \\
&\rightarrow \texttt{duration\_node}.
\end{aligned}
\end{equation}

This ordering is deliberate: location grounding provides the spatial context required for travel-time propagation and duration scheduling, while each module reads and writes well-scoped state fields, preventing tool-call drift and ensuring reproducible long-horizon generation.

\subsubsection{Location Grounding 
}
\label{subsec:location}
For each activity visit $a_i$, we ground a location in two steps: (1) construct a feasible candidate POI set $P(a_i)$, and (2) score candidates by combining their likelihood under user/similar-user preferences with distance-compatibility, then sample accordingly.
\paragraph{Candidate construction with feasible set and controlled exploration}
For each activity $a_i$, we \emph{retrieve} a candidate POI set rather than sampling from the full location space.
We first build an individual-specific feasible set $\mathcal{P}_{u}(a_i)$ from historical visits.
To enable controlled exploration beyond personal history, we augment candidates using a top-$K$ similar-individual pool obtained via similarity matching over mobility signatures (e.g., spatial scale, temporal rhythm, and activity/transition distributions, optionally with co-location signals).

This yields an augmented location memory $\mathcal{P}_{\text{sim}}(a_i)$ from similar individuals, and we take $\mathcal{P}(a_i)=\mathcal{P}_{u}(a_i)\cup\mathcal{P}_{\text{sim}}(a_i)$.
If $\mathcal{P}(a_i)$ is empty, we emit an explicit invalid marker (rather than silently sampling) to avoid cascading errors.

\paragraph{Distance-aware scoring and stochastic selection} To choose a plausible next POI, we score each candidate by combining likelihood (how likely the user or similar users visits it for this activity) with distance based feasibility (how consistent the travel distance is with the user’s typical transitions).
For each candidate $p\in\mathcal{P}(a_i)$, we compute a composite score:

\vspace{-3mm}

\begin{equation}
S(p)=\lambda_f\cdot s_{\text{freq}}(p) + \lambda_d\cdot s_{\text{dist}}(p),
\end{equation}

where the frequency prior mixes individual and neighbor preferences with exploration gate $\alpha$:

\vspace{-3mm}

\begin{equation}
s_{\text{freq}}(p)=(1-\alpha)\,P_u(p\mid a_i)+\alpha\,P_{\text{sim}}(p\mid a_i).
\end{equation}
The exploration gate $\alpha$ balances individual fidelity and controlled diversity. Moderate changes in $\alpha$ have limited impact on our statistical or anomaly metrics, but $\alpha=0$ can yield near-copy trajectories with lower downstream utility, while excessive peer weighting may drift beyond the target individual's mobility scope.

To enforce physically plausible transitions, distance compatibility is measured against the previous grounded POI $p_{i-1}$ and the individual's historical transition-distance regime for $(a_{i-1},a_i)$:

\vspace{-6mm}

\begin{equation}
s_{\text{dist}}(p)=\exp\!\left(-\beta\cdot \left| \mathrm{dist}(p_{i-1},p)-\bar{\ell}_{u}(a_{i-1},a_i)\right|\right),
\end{equation}
where $\mathrm{dist}(\cdot,\cdot)$ is the Haversine distance and $\bar{\ell}_u(a_{i-1},a_i)$ is the individual-specific mean transition distance (with robust defaults if unavailable). Finally, we sample $p_i$ from the normalized $\{S(p)\}$, yielding stochastic yet profile-constrained location grounding with built-in personalization and controlled exploration, without any model fine-tuning.

\subsubsection{Travel Time
}
\label{subsec:time}

Given the grounded POIs $\{p_i\}_{i=1}^{N_{u,d}}$ and activities $\{a_i\}_{i=1}^{N_{u,d}}$, the temporal worker constructs an \emph{irregular} daily timeline by iteratively producing each visit's start time $t_i^{s}$ and duration $\delta_i$ (thus end time $t_i^{e}$).
Time is treated as a first-class continuous variable, avoiding the granularity loss from fixed-bin discretization and enabling explicit feasibility control through budget and kinematic priors.

\paragraph{Cold-start initialization for the first visit}
For $i=1$, we initialize $t_1^{s}$ from an individual- and weekday-conditioned prior of the \emph{first-visit start time} estimated from history. If unavailable, we fall back to a conservative default (e.g., morning start) and add a small random offset to avoid degenerate identical schedules across days.

\paragraph{Distance- and kinematics-aware travel-time propagation}
For $i>1$, the workflow advances time using a travel-time estimate driven by (i) the geographic distance between consecutive POIs and (ii) an individual-specific \emph{kinematic prior} captured by historical transition speeds:
\begin{equation}
\begin{aligned}
t_i^{s} &= t_{i-1}^{e} + \Delta t_i^{\text{travel}}, \\
\Delta t_i^{\text{travel}} &=
\mathrm{clip}\!\left(
\frac{dist(p_{i-1},p_i)}{v_u(a_{i-1},a_i)}\cdot 60,\;
\Delta_{\min},\;\Delta_{\max}
\right).
\end{aligned}
\end{equation}
Here $dist(\cdot,\cdot)$ is the Haversine distance, $v_u(a_{i-1},a_i)$ is the historical mean speed for transition $(a_{i-1}\!\rightarrow\!a_i)$, and $\mathrm{clip}(\cdot)$ enforces pre-defined time feasibility bounds (e.g., $\Delta_{\min}=5$ min, $\Delta_{\max}=180$ min).
When $v_u$ is missing or unreliable, we use robust defaults based on distance regime (e.g., walk vs.\ drive), preserving kinematic plausibility without requiring the LLM to reason over high-dimensional mobility dynamics.

\subsubsection{Duration estimation}
Unlike travel time, activity duration is highly context-dependent (e.g., \texttt{Work} vs.\ \texttt{EatOut}) and may adapt under schedule pressure.
We therefore delegate duration estimation to an LLM worker that conditions on retrieved evidence and the current generation state:
\begin{itemize}
    \item current start time $t_i^{s}$ and remaining daily time budget,
    \item current activity $a_i$ and its historical duration prior,
    \item remaining activities $[a_{i+1},\ldots,a_{N_{u,d}}]$ and their expected total time,
    \item optional individual-specific duration tendencies.
\end{itemize}
The LLM outputs a strict JSON object (e.g., \texttt{\{"duration\_minutes": 45\}}).
A lightweight verifier enforces schema validity and feasibility:
$
\delta_i \in 
[\delta_{\min},\; \min(\delta_{\max},\ \text{budget\_left})],
$
with retry-on-violation and deterministic fallback to the historical prior if parsing or validation fails. This evidence-to-decision design preserves semantic flexibility in duration choices while guaranteeing coherent and time-feasible execution. Finally, we update the end time as
$
t_i^{e} = t_i^{s} + \delta_i,
$
and proceed to the next visit, cumulatively generating a full-day irregular timeline that is semantically coherent and physically feasible.

\section{Experiments}

In this section, we evaluate \methodname for human mobility trajectory generation on two large-scale synthetic datasets. We compare it against two prevailing paradigms: (i) trajectory-level fine-tuned LLM generators, represented by \textbf{Geo-Llama} \cite{li2025geo}, and (ii) state-of-the-art non-LLM neural generators (see Section \ref{sec:baselines}). 
This setup tests whether a zero-shot, workflow-managed agent can match or surpass both fine-tuning-based LLM generators and state-of-the-art non-LLM continuous-time neural baselines in spatiotemporal fidelity and semantic coherence, while avoiding costly parameter updates.

Beyond conventional aggregate spatiotemporal statistics, we introduce an anomaly detection-based evaluation framework to probe semantic coherence and behavioral plausibility. Specifically, we incorporate two complementary anomaly detectors, \textbf{ICAD} \cite{azarijoo2025icad} and \textbf{BeSTAD} \cite{xie2025bestad}, which capture different abnormal patterns and provide diagnostic signals that aggregation-level metrics can miss. Overall, our experiments ask whether \methodname (1) preserves basic spatiotemporal statistics fidelity, (2) improves semantic and behavioral plausibility under anomaly-based scrutiny, and (3) achieves these gains with lower computational overhead by avoiding fine-tuning. Our code can be accessed at: 
\texttt{\url{https://github.com/Emory-AIMS/TrajGenAgent}}.


\begin{table}[ht]
    \centering
    
    \label{tab:dataset_statistics}
    \begin{tabular}{l|c|c}
        \hline
        \textbf{Statistic} & \textbf{NumoSim} & \textbf{MobilitySyn} \\ \hline
        Total daily trajectories used & 34,000 & 34,000 \\
        Avg.\ stay points per trajectory & 7.2 & 8.7 \\
        \# Individuals & 1,200 & 1,200 \\
        \# Activity types & 16 & 6 \\
        \hline
    \end{tabular}
    \caption{Training Dataset Statistics}
    \vspace{-4mm}
\end{table}


\subsection{Datasets \& Preprocessing}

\indent We conduct experiments on two synthetic mobility datasets: the open-source benchmark \textbf{NumoSim} and our simulated \textbf{MobilitySyn} dataset.

\begin{itemize}
    \item \textbf{NumoSim.} NumoSim \cite{stanford2024numosim} is a large-scale synthetic mobility benchmark for anomaly detection, providing 8 weeks of stay-point trajectories for 200{,}000 individuals in Los Angeles.

    \item \textbf{MobilitySyn.} We create MobilitySyn by following the simulation framework in \cite{yoginath2025scalable} to generate a realistic week-long mobility trace for 5{,}000 individuals over a metropolitan area. The simulator produces second-by-second GPS records, which we convert into visit-wise (stay-point) trajectories for evaluation.
\end{itemize}

\noindent \textbf{Trajectory Representations}.  
 Our baselines cover two trajectory representations: (i) \emph{fixed-interval} sequences with 96 steps per day (15-minute bins), and (ii) \emph{visit-wise} sequences with variable length. \methodname, Geo-Llama \cite{li2025geo}, and Geo-CETRA \cite{lin2024controllable} (see details in Section \ref{sec:baselines}) operate on visit-wise trajectories. Geo-Llama represents each visit by a POI ID and a discretized timestamp, whereas \methodname uses POI IDs with continuous timestamps and an intermediate activity type. Geo-CETRA generates continuous locations and times; for fair comparison, we discretize its outputs to the same evaluation grid/time bins. Specifically, we use 15-minute intervals and grid sizes of 0.5\,km for NumoSim and 0.7\,km for MobilitySyn.

\vspace{-1mm}

\subsection{Baselines}
\label{sec:baselines}
\indent We evaluate the performance of our model against the following six state-of-the-art baselines:
\begin{itemize}
    \item  \textbf{GRU}~\cite{lecun1998gradient} and \textbf{LSTM}~\cite{hochreiter1997long}: Recurrent neural networks that are efficient for sequential data generation. These models are able to predict the next location based on historically visited locations.

    \item \textbf{Transformer} \cite{vaswani2017attention}: A powerful deep learning model used in various natural language processing (NLP) and computer vision tasks that leverages self-attention mechanisms. A multi-layer Transformer decoder is utilized for trajectory generation.
    \item  \textbf{SeqGAN} \cite{yu2017seqgan}: A sequence GAN that introduces a discriminator as a reward signal to guide the gradient policy update of the generator, which performs the next location prediction task based on the past states.

    \item \textbf{Geo-CETRA} \cite{lin2024controllable}: A spatiotemporal point process-based framework for trajectory generation that incorporates constraint factorization and beam decoding to produce realistic trajectories.

    \item \textbf{Geo-Llama} \cite{li2025geo}: An LLM-based generator that encodes daily trajectories as structured visit sequences and learns spatiotemporal dependencies via parameter-efficient fine-tuning (e.g., LoRA). It applies visit-wise permutation to encourage learning temporal regularities from visit-level time attributes rather than sequence order. 

\end{itemize}

\begin{table*}[htb]
    \centering
    {
    \begin{tabular}{c|l|ccccc|cc}
        \multirow{2}{*}{\textbf{Dataset}} & \multirow{2}{*}{\textbf{Model}} & \multicolumn{5}{|c|}{\textbf{Trajectory-level} ($\downarrow$)} & \multicolumn{2}{c}{\textbf{Global-level} ($\downarrow$)}\\
        & & \textbf{Distance} & \textbf{G-radius} & \textbf{Duration} & \textbf{DailyLoc} & \textbf{I-rank} & \textbf{G-rank} & \textbf{Transition}\\ \hline
        \multirow{7}{*}{\textit{NumoSim}}
                & GRU         & 0.0111 & 0.2557 & 0.2145 & 0.1561 & 0.0137 & 0.0159 & 0.0156 \\
                & LSTM        & 0.0146 & 0.3113 & 0.3013 & 0.1981 & 0.0893 & 0.0134 & 0.0150 \\
                & Transformer & 0.0082 & 0.2945 & 0.2150 & 0.1620 & 0.0079 & 0.0112 & 0.0118 \\
                & SeqGAN      & 0.0085 & 0.0998 & 0.2410 & 0.1585 & 0.0082 & 0.0107 & 0.0120 \\
                & Geo-CETRA   & 0.0093 & 0.3337 & 0.0060 & 0.1128 & 0.0002 & 0.0002 & 0.0088 \\
                & Geo-Llama   & 0.0075 & 0.2361 & \textbf{0.0028} & \textbf{0.0128} & \textbf{0.0001} & \textbf{0.0001} & 0.0087 \\
                & \textbf{\methodname}  & \textbf{0.0006} & \textbf{0.0993} & 0.0155 & 0.2117 & 0.0002 & 0.0002 & \textbf{0.0075} \\ \hline
        \multirow{7}{*}{\textit{MobilitySyn}} 
                & GRU         & 0.0116 & 0.1859 & 0.1747 & 0.3368 & 0.0082 & 0.0097 & 0.0132 \\       
                & LSTM        & 0.0131 & 0.2823 & 0.1680 & 0.3046 & 0.0044 & 0.0078 & 0.0135 \\
                & Transformer & 0.0085 & 0.3760 & 0.1510 & 0.2810 & 0.0047 & 0.0065 & 0.0115 \\
                & SeqGAN      & 0.0089 & 0.0738 & 0.1344 & 0.2437 & 0.0035 & 0.0062 & 0.0108 \\
                & Geo-CETRA   & 0.0276 & 0.5784 & 0.0319 & 0.1573 & 0.0006 & 0.0006 & 0.0083 \\
                & Geo-Llama   & 0.0268 & 0.5528 & \textbf{0.0241} & 0.1209 & 0.0005 & 0.0005 & 0.0078 \\
                & \textbf{\methodname}  & \textbf{0.0000} & \textbf{0.0051} & 0.1308 & \textbf{0.0000} & \textbf{0.0003} & \textbf{0.0003} & \textbf{0.0000} \\ \hline
    \end{tabular}}
    \caption{Aggregation-level spatiotemporal statistical metrics of the trajectory generation.}
    \label{tab:uncond-exp}
\end{table*}



\subsection{Evaluation Metrics}

Our primary intended use case is providing synthetic trajectory data for counterfactual analysis of urban mobility patterns, rather than optimizing a specific downstream task such as next-POI recommendation. We therefore evaluate generated trajectories from two complementary perspectives: (1) \emph{spatiotemporal statistics fidelity} under aggregated mobility statistics, and (2) \emph{behavioral semantics} under downstream anomaly detection models. 

\subsubsection{Aggregation-level Spatiotemporal Statistics Metrics}

Following prior mobility-generation evaluations \cite{feng2020learning,zhang2023csgan}, we compare generated and real trajectories through population- and trajectory-level mobility distributions.

\textbf{Distance} is the distribution of cumulative travel distance per user per day.
\textbf{G-radius} (radius of gyration) measures the distribution of daily spatial movement range.
\textbf{Duration} is the distribution of dwell time per visited location.
\textbf{DailyLoc} is the distribution of the number of visited locations per user per day.
\textbf{G-rank} is the global visit-frequency distribution over top-100 visited locations.
\textbf{I-rank} is the per-user counterpart of G-rank.

For these distributional properties, we compute Jensen--Shannon divergence (JSD): $\text{JSD}\left(\mathbb{D}, \mathbb{D}'\right) = h\left(\frac{\mathbb{D} + \mathbb{D}'}{2}\right) - \frac{h(\mathbb{D}) + h(\mathbb{D}')}{2}$ where $\mathbb{D}$ and $\mathbb{D}'$ denote real and generated distributions, and $h(\cdot)$ is Shannon entropy.
Lower JSD indicates better agreement with real mobility statistics.

We additionally compare transition dynamics.
\textbf{Transition} is the location-to-location transition probability matrix over discretized locations $\mathcal{G}$.
Its discrepancy is measured by Frobenius norm: $\|P_\mathbb{D} - P_\mathbb{D'}\|_F = \sqrt{\sum_{l_1=1}^{|\mathcal{G}|} \sum_{l_2=1}^{|\mathcal{G}|} \left| P_\mathbb{D}(l_1, l_2) - P_\mathbb{D'}(l_1, l_2) \right|^2}$. Lower values indicate better preservation of transition structure. We note that G-rank and Transition correspond to global-level while others correspond to trajectory-level patterns.

\subsubsection{Anomaly-Detection Evaluation}

Aggregate spatiotemporal fidelity can be achieved even when trajectories are behaviorally implausible at the visit or individual level. To assess behavioral realism beyond population-level metrics, we evaluate generated trajectories with two complementary anomaly detectors: \textbf{ICAD} and \textbf{BeSTAD}. Since human mobility realism is inherently multi-faceted and partly subjective, we use these detectors as behavior-aware diagnostic proxies rather than exhaustive measures of semantic coherence. Intuitively, trajectories that are both statistically realistic and semantically coherent in individual behavior should be \emph{harder} for these detectors to distinguish from trajectories in the training set.


\paragraph{ICAD (visit-level multi-context detector)}
ICAD is a self-supervised autoregressive framework that decomposes each visit into \emph{location}, \emph{arrival time}, and \emph{departure time}, and learns next-visit regularities under normal mobility patterns \cite{azarijoo2025icad}. 
For anomaly scoring, it uses top-$k$ deviation for discrete spatial prediction and a mode-margin density score (GMM-based) for continuous temporal components, then fuses component-wise deviations into a final anomaly score.
A key property is interpretability: ICAD can attribute abnormality to spatial, temporal, or compound deviations.
This makes ICAD suitable for testing whether generated visits preserve fine-grained spatiotemporal consistency.

\paragraph{BeSTAD (individual-level behavioral shift detector)}
BeSTAD targets \emph{individual-level} anomalies by modeling individualized behavior clusters in a past ``normal'' period and comparing cluster alignment in a future period \cite{xie2025bestad}. 
It integrates temporal behavior signals with multi-scale spatial semantics (including point/line/polygon context, e.g., from OpenStreetMap (OSM), via the Hexagonal Hierarchical Geospatial Indexing System (H3)-based indexing) to capture richer behavioral context.
Its anomaly score emphasizes cross-period behavioral shifts and emerging routine changes rather than isolated single-visit outliers.
Therefore, BeSTAD complements ICAD by stressing high-level behavioral coherence at individual level.

\paragraph{Metrics}
We report anomaly-detection outputs, AUROC and average precision (AP), by applying detectors with identical settings to distinguish real trajectories (normal) from generated ones (treated as anomalies). Unless otherwise specified, we use a balanced split with equal positive and negative samples, for which random guessing yields AUROC $\approx 0.5$ and AP $\approx 0.5$. This conservative setting avoids inflated performance due to class imbalance. From a generation-quality perspective, scores closer to  chance  indicate lower separability---i.e., the generated data is harder to distinguish from real mobility data under the detector---and therefore exhibit higher behavioral semantic fidelity. 

\vspace{-1mm}
\subsection{Implementation Details}

\begin{itemize}
    \item \textbf{GRU, LSTM, and Transformer} 
    are trained for 200 epochs using the Adam optimizer with a learning rate of 0.001. Models share an embedding size of 256, with GRU and LSTM using 6 layers of 512 hidden units, while Transformer adopts a decoder-only architecture with 4 layers and 4 attention heads.

    \item \textbf{SeqGAN}
    includes an LSTM-based generator trained with 16-dimensional embeddings and 16 hidden units. Discriminator employs diverse filter sizes and counts, and rollout number of 8. The entire pipeline carries 40 epochs of pre-training and 20 epochs of adversarial training.

    \item \textbf{Geo-CETRA}
    employs a conditional Gaussian Mixture Model  with 8 spatial and temporal mixture components and a beam search strategy with a beam size 10 and top $k$=3. Optimization is performed with the Adam optimizer, a learning rate of 0.01, a scheduler with a decay factor of 0.99, and z-score normalization for input data.
    
    \item \textbf{Geo-Llama} 
    fine-tunes the \texttt{Llama-2-7b-chat-hf} model using LoRA with a batch size of 48, a learning rate of 0.00001, LoRA alpha32, LoRA dropout 0.02, LoRA r 16, and 20 epochs. Sampling uses temperature 1.2. 

    \item \textbf{\methodname}
    performs zero-shot generation without model fine-tuning, using \texttt{Qwen2.5-32B-Instruct} as the backbone LLM. Both workflow stages are served with \texttt{vLLM} for high-throughput inference (temperature $0.90$, top-$p$ $0.95$, max tokens $1024$, max context length $8192$).

\end{itemize}

\subsection{Aggregation-level Spatiotemporal Statistics}

Table~\ref{tab:uncond-exp} reports the trajectory generation performance under aggregation-level spatiotemporal statistical metrics. Overall, \methodname achieves the strongest \emph{spatial} alignment across both datasets: it consistently attains the lowest divergence on distance-based metrics (\textbf{Distance}, \textbf{G-radius}) and improves global transition realism (\textbf{Transition}), highlighting the benefit of inference-time spatiotemporal grounding with peer-augmented individualized evidence and distance-/kinematics-aware priors. On \textit{MobilitySyn}, \methodname is near-perfect on most location- and transition-related metrics (Distance, G-radius, DailyLoc, I-rank/G-rank, and Transition), indicating that the workflow can robustly reconstruct visit-wise spatial statistics and movement structure without parameter updates. On \textit{NumoSim}, \methodname remains highly competitive on spatial and transition metrics, while Geo-Llama is stronger on time-centric statistics (notably \textbf{Duration} and \textbf{DailyLoc}). This gap likely arises because duration grounding is sensitive to accumulated state errors and conflicts between the sampled activity scaffold, remaining time budget, and sparse individual temporal evidence. We attribute this to the increased behavioral complexity of NumoSim (more activity types and richer daily schedules), where fine-tuning on structured visit tokens can more precisely fit dwell-time and visit-count distributions; in contrast, our budget-aware duration module prioritizes schedule feasibility and semantic plausibility; since it is not directly optimized to match aggregate duration distributions, it can be less calibrated on micro-level duration statistics on more behaviorally complex datasets. A key factor across baselines is trajectory representation.
The fixed-interval generators (GRU/LSTM/Transformer/SeqGAN) operate on 96-step sequences per day with implicit time encoded through discretized bins, which produces long sequences with repeated states (stay points) and can obscure fine-grained temporal patterns. In contrast, visit-wise generators (Geo-CETRA, Geo-Llama, and \methodname) model daily trajectories as variable-length visit sequences, which better matches the underlying event structure and generally improves transition- and rank-related statistics. Among the fixed-interval baselines, GRU/LSTM/Transformer capture some location-frequency patterns reasonably well, whereas adversarial training (SeqGAN) is more sensitive to optimization instability and can degrade fidelity on both spatial and temporal metrics.

\subsection{Anomaly-Detection Evaluation}

Table~\ref{tab:anomaly_combined} reports anomaly-detection results under a balanced generated-vs-real split, where AUROC/AP closer to $0.5$ indicates \emph{lower separability} and thus better semantic coherence. Since ICAD is prediction-based, it naturally supports evaluation at both the visit-level and the individual-level, whereas BeSTAD is designed for individual-level behavioral shift detection and is therefore reported at the individual-level only. Because anomaly detectors operate on POIs and activity types, we focus on the three strongest visit-wise trajectory generators without location grids, which means they can be directly mapped to POI/GPS coordinates without introducing coarse grid inversion artifacts: Geo-CETRA, Geo-Llama, and \methodname.

On \textit{NumoSim}, which has richer underlying spatiotemporal dynamics and a stronger activity type diversity, \methodname consistently yields AUROC/AP values closest to $0.5$ across both detectors, indicating that its generated trajectories are the hardest to distinguish from real ones. Geo-Llama is generally stronger than Geo-CETRA, but both remain substantially more separable than \methodname under ICAD's fine-grained spatiotemporal scrutiny and BeSTAD's individual-level behavioral shift detection. These results align with our design goal: inference-time spatiotemporal grounding with individualized evidence and kinematics-aware priors can preserve semantic plausibility while avoiding systematic artifacts detectable by anomaly models. 

On \textit{MobilitySyn}, which has simpler underlying mobility patterns and fewer activity types, results are more mixed. Geo-Llama is closest to chance under BeSTAD, while Geo-CETRA attains the best (closest-to-$0.5$) ICAD visit-level scores; \methodname remains competitive and achieves the strongest ICAD individual-level AP, suggesting improved individual-level behavioral stability under ICAD. We conjecture that on simpler simulation dynamics, direct fine-tuning on structured tokens (Geo-Llama) or end-to-end neural approach with rule-based decomposition of visit-wise movement constraints as priors (Geo-CETRA) can already obscure many detector cues, while the strength of \methodname in evidence-driven semantic planning and kinematics-aware grounding are more pronounced on richer, behaviorally more diverse datasets.

\begin{table}[htb]
    \centering
    \setlength{\tabcolsep}{3pt} 
    \scalebox{0.88}{
    \begin{tabular}{c|l|c|cc|cc}
       \multirow{3}{*}{\textbf{Dataset}} & \multirow{3}{*}{\textbf{Model}} & \textbf{BeSTAD} & \multicolumn{2}{c|}{\textbf{ICAD (Visit-level)}} & \multicolumn{2}{c}{\textbf{ICAD (Individual-level)}} \\
       & & ($\to 0.5$) & \multicolumn{2}{c|}{($\to 0.5$)} & \multicolumn{2}{c}{($\to 0.5$)} \\
       & & \textbf{AUROC} & \textbf{AP} & \textbf{AUROC} & \textbf{AP} & \textbf{AUROC} \\ \hline
       \multirow{3}{*}{\textit{NumoSim}}
                & Geo-CETRA   & 0.5767 & 0.7234 & 0.6414 & 0.7806 & 0.8821 \\
                & Geo-Llama   & 0.3375 & 0.7046 & 0.6057 & 0.7888 & 0.8962 \\
                & \textbf{\methodname} & \textbf{0.5008} & \textbf{0.5255} & \textbf{0.5368} & \textbf{0.5690} & \textbf{0.6398} \\ \hline
       \multirow{3}{*}{\textit{MobilitySyn}} 
                & Geo-CETRA   & 0.8192 & \textbf{0.4668} & \textbf{0.5435} & 0.7934 & 0.7314 \\
                & Geo-Llama   & \textbf{0.5025} & 0.5885 & 0.5969 & 0.7283 & \textbf{0.6629} \\
                & \textbf{\methodname} & 0.6817 & 0.6318 & 0.6761 & \textbf{0.6342} & 0.7194 \\ \hline
    \end{tabular}}
    \caption{Trajectory generation performance under BeSTAD and ICAD anomaly detection. With balanced split (pos/neg=0.5), AP and AUROC closer to 0.5 is better.}
    \label{tab:anomaly_combined}
    \vspace{-2.5mm}
\end{table}
\subsection{Computational Efficiency}

Table~\ref{tab:gpu_hours} summarizes the end-to-end computational cost under a unified setting, with all timings measured on a single NVIDIA H100 GPU with maximized memory utilization. For parameter-updated baselines, the reported cost includes training plus inference; for TrajGenAgent, which performs zero-shot generation without parameter updates, it includes only inference-time generation. In both cases, the workload uses the same scale: 34{,}000 historical daily trajectories from 1{,}200 individuals are used as the training set or historical evidence, and each method generates 34{,}000 daily trajectories.

Overall, \methodname offers a strong cost--quality trade-off: it matches or slightly exceeds Geo-Llama in generation quality while avoiding costly parameter updates by injecting fine-grained spatiotemporal evidence at inference time through a deterministic, verifier-guarded workflow. Such transferability is particularly valuable for transfer to new cities or mobility regimes, where patterns shift and repeated fine-tuning is impractical under limited budgets. Among advanced baselines, Geo-CETRA is relatively efficient by combining rule-based decomposition of visit-wise movement constraints with an efficient Transformer backbone, yielding moderate GPU hours with competitive quality. In contrast, Geo-Llama attains strong spatiotemporal fidelity via fine-tuning-driven sequence modeling over structured tokens, incurring the highest compute among the strong baselines. SeqGAN is the least cost-effective: adversarial training and Monte-Carlo rollout-based reward feedback dominate runtime, leading to very high GPU hours despite a lightweight generator and weaker performance. For naive baselines, GRU/LSTM are the most compute-efficient but deliver weaker fidelity, while the vanilla Transformer improves fidelity at a higher training cost, reflecting the standard capacity--efficiency trade-off. Taken together, these results highlight that \methodname achieves comparable or better quality than fine-tuned LLMs with substantially lower end-to-end compute, narrowing the apparent gap between low compute cost and high generation fidelity.

\begin{table}[htb]
    \centering
    \renewcommand{\arraystretch}{1.1} 
    \resizebox{\linewidth}{!}{
    \begin{tabular}{l|cccc}
        \hline
        \textbf{Model} & GRU & \textbf{\methodname} & LSTM & Transformer \\ \hline
        \textbf{GPU Hours} ($\downarrow$) & \textbf{1.25} & 1.67 & 1.83 & 3.17 \\ 
        \hline
        \multicolumn{5}{c}{} \\[-0.8em] 
        \hline
        \textbf{Model} & Geo-CETRA & SeqGAN & Geo-Llama & \\ \hline
        \textbf{GPU Hours} ($\downarrow$) & 3.38 & 20.62 & 24.77 & \\ 
        \hline
    \end{tabular}
    }
    \caption{Average computational cost per dataset on a single NVIDIA H100 GPU, including generating 34{,}000 daily trajectories and training with the same number if required.}
    \label{tab:gpu_hours}
    \vspace{-3mm}
\end{table}

\subsection{Tool-Invocation Stability: Deterministic Workflow vs. Free-form Tool Calling}
\label{sec:free_form_tool_calling}

We further evaluate tool-invocation stability with a simplified spatiotemporal grounding setting to justify our use of deterministic workflow execution instead of zero-shot free-form tool calling. Specifically, we fix the activity chains and evaluate only Stage-2 spatiotemporal grounding, where each daily trajectory contains seven activities. This grounding process is sequentially dependent: the POI selected for the current visit affects travel-time estimation, the resulting end time determines the next visit's start time, and both variables condition subsequent POI and time decisions. Therefore, a single missing or malformed tool call can break downstream state variables and cause cascading failures. Both variants use the same \texttt{Qwen2.5-32B-Instruct} backbone and a simplified two-tool interface, \texttt{location} and \texttt{time}, which gives the free-form variant a favorable setting by minimizing tool-selection complexity. The free-form variant receives the activity chain, tool schemas, and a semantic instruction to convert the chain into a complete trajectory, and the LLM autonomously decides when and how to invoke tools. In contrast, our workflow-managed variant executes the same grounding task under a fixed state-machine order. Although tool-call fine-tuning with scenario--tool--JSON traces can improve autonomous invocation, it introduces additional data and training costs and may require a smaller backbone under the same hardware budget. We therefore compare against zero-shot free-form tool calling using the same backbone without fine-tuning.


To avoid early termination after a failed call, the tools include default-value fallbacks that allow the grounding process to continue. These fallbacks only keep later visits executable, but may still introduce inaccurate state variables that propagate through subsequent grounding steps. Table~\ref{tab:free_form_tool_calling} therefore reports invocation success under this fallback-enabled execution setting. Trajectory-level success requires that all visits in a daily trajectory complete the required location and time calls without fallback-induced substitution. Visit-level success measures the fraction of visits in which the required calls are completed. Even under this simplified setting, zero-shot free-form tool calling falls short of a practically usable level of tool-invocation stability for sequential grounding, whereas deterministic workflow achieves perfect invocation with explicitly predefined control flow and state dependencies.

\begin{table}[t]
\centering
\small
\scalebox{0.9}{%
\begin{tabular}{lcc}
\toprule
\textbf{Tool-calling strategy} 
& \textbf{Traj.-level success} $\uparrow$
& \textbf{Visit-level success} $\uparrow$ \\
\midrule
Free-form tool calling 
& 9.8\% 
& 59.3\% \\
Deterministic workflow 
& 100.0\% 
& 100.0\% \\
\bottomrule
\end{tabular}%
}
\caption{Tool-invocation stability comparison between free-form tool calling and deterministic workflow execution under fallback-enabled setting.}
\label{tab:free_form_tool_calling}
\vspace{-4mm}
\end{table}

\begin{table}[htb]
    \centering
    {
    \begin{tabular}{l|cc|cc}
        \multirow{2}{*}{\textbf{Metric} ($\downarrow$)} & \multicolumn{2}{c|}{\textbf{NumoSim}} & \multicolumn{2}{c}{\textbf{MobilitySyn}} \\
        & \textbf{\methodname} & \textbf{\textit{w/o kin.}} & \textbf{\methodname} & \textbf{\textit{w/o kin.}} \\ \hline
        \textbf{Distance}   & \textbf{0.0006} & 0.0028 & \textbf{0.0000} & \textbf{0.0000} \\
        \textbf{G-radius}   & \textbf{0.0993} & 0.1508 & 0.0051 & \textbf{0.0004} \\
        \textbf{Duration}   & \textbf{0.0155} & 0.0198 & \textbf{0.1308} & 0.3732 \\
        \textbf{DailyLoc}   & \textbf{0.2117} & 0.2476 & \textbf{0.0000} & \textbf{0.0000} \\
        \textbf{I-rank}     & \textbf{0.0002} & 0.0006 & 0.0003 & \textbf{0.0001} \\
        \textbf{G-rank}     & \textbf{0.0002} & 0.0006 & 0.0003 & \textbf{0.0001} \\
        \textbf{Transition} & \textbf{0.0075} & 0.0077 & \textbf{0.0000} & \textbf{0.0000} \\ \hline
    \end{tabular}}
    \caption{Ablation study of \methodname on kinematics design with aggregation-level spatiotemporal statistical metrics.}
    \label{tab:ablation}
\end{table}

\begin{table}[htb]
    \centering
    \scalebox{0.9}{
    \begin{tabular}{l|cc|cc}
        \multirow{2}{*}{\textbf{Metric}} & \multicolumn{2}{c|}{\textbf{\textit{NumoSim}}} & \multicolumn{2}{c}{\textbf{\textit{MobilitySyn}}} \\
        & \textbf{\methodname} & \textbf{\textit{w/o kin.}} & \textbf{\methodname} & \textbf{\textit{w/o kin.}} \\ \hline
        \multicolumn{5}{c}{\textbf{BeSTAD}} \\ \hline
        AUROC ($\to 0.5$)       & \textbf{0.5008} & 0.5104 & \textbf{0.6817} & 0.7400 \\ \hline
        \multicolumn{5}{c}{\textbf{ICAD (Visit-level)}} \\ \hline
        AP ($\to 0.5$)          & \textbf{0.5255} & 0.6692 & \textbf{0.6318} & 0.7143 \\
        AUROC ($\to 0.5$)       & \textbf{0.5368} & 0.6483 & \textbf{0.6761} & 0.6827 \\ \hline
        \multicolumn{5}{c}{\textbf{ICAD (Agent-level)}} \\ \hline
        AP ($\to 0.5$)          & \textbf{0.5690} & 0.9034 & \textbf{0.6342} & 0.9922 \\
        AUROC ($\to 0.5$)       & \textbf{0.6398} & 0.9143 & \textbf{0.7194} & 0.9925 \\ \hline
    \end{tabular}}
    \caption{The Kinematics Ablation study of \methodname under BeSTAD and ICAD anomaly detection.}
    \label{tab:ablation_anomaly}

\end{table}
\subsection{Ablation Study: Kinematics-aware Grounding}
Tables~\ref{tab:ablation} and \ref{tab:ablation_anomaly} show that the kinematics-aware module is a key contributor to \methodname: removing it degrades mobility pattern fidelity especially under the scrutiny of ICAD/BeSTAD. This confirms that physics-/kinematics-informed travel-time propagation helps maintain coherent temporal progression and behavior-level plausibility beyond what aggregate metrics alone reveal. More broadly, the ablation highlights a core advantage of our workflow-managed agent design: specialized constraints and domain modules can be plugged in or removed flexibly, which is challenging for end-to-end approaches.

\vspace{-1.5mm}
\subsection{Parameter Study: LLM Sampling Temperature Impacts}

\begin{table}[t]
    \centering
    \scriptsize
    \setlength{\tabcolsep}{2.0pt}
    \renewcommand{\arraystretch}{1.05}
    \resizebox{\columnwidth}{!}{
    \begin{tabular}{l|ccc|ccc|ccc}
        \hline
        ~
        & \multicolumn{3}{c|}{\textbf{$T=0.5$}} 
        & \multicolumn{3}{c|}{\textbf{$T=0.9$}} 
        & \multicolumn{3}{c}{\textbf{$T=1.5$}} \\
        ~
        & \textbf{Total$\downarrow$} & \textbf{Schema$\downarrow$} & \textbf{Constr.$\downarrow$}
        & \textbf{Total$\downarrow$} & \textbf{Schema$\downarrow$} & \textbf{Constr.$\downarrow$}
        & \textbf{Total$\downarrow$} & \textbf{Schema$\downarrow$} & \textbf{Constr.$\downarrow$} \\
        \hline
        \textbf{Avg.\ FR(\%)} & 14.3\% & \textbf{1.8\%} & 12.5\% 
        & \textbf{9.1\%} & 3.6\% & \textbf{5.5\%} 
        & 28.3\% & 15.2\% & 13.1\% \\
        \hline
    \end{tabular}}
    \caption{LLM verifier-triggered failure rates (FR) under different sampling temperatures $T$ (averaged across datasets).}
    \label{tab:temp_verifier}
    \vspace{-5mm}
\end{table}

Table~\ref{tab:temp_verifier} reports verifier-triggered failure rates under three sampling temperatures (averaged across datasets). We mark a daily run as failed if either Stage~1 (activity chain) or Stage~2 (duration) LLM output violates the required schema or basic feasibility constraints (e.g., chain length/duration bounds).

Schema failures increase monotonically with temperature, as higher $T$ flattens the token distribution and destabilizes structured outputs. Constraint failures are non-monotonic: very low $T$ is mode-seeking and can repeat typical but infeasible values under budget/bound checks, while very high $T$ induces out-of-bound or implausible samples; $T=0.9$ provides the best balance and the lowest overall failure rate.

\balance
\vspace{0mm}
\section{Conclusion}
We presented \methodname, a zero-shot, semantic-aware hierarchical LLM-agent framework for human mobility trajectory generation. It couples an orchestrator LLM for individual- and weekday-conditioned activity chain planning with a deterministic, verifier-guarded worker workflow for kinematics-aware spatiotemporal grounding via inference-time tool integration (without fine-tuning or parameter updates). Across both  large-scale simulation datasets, \methodname\ consistently outperforms baselines under complementary evaluations, including traditional spatiotemporal statistical metrics and our anomaly-detection-based semantic-aware metrics (ICAD/BeSTAD). 

Despite the promising results, several limitations suggest directions for future work. Our temporal grounding relies on retrieval-driven priors and bounded constraints; even augmented with LLMs, achieving highly accurate, individual-specific temporal modeling remains challenging and motivates stronger constraint-aware neural temporal modules. In addition, our verifiers mainly enforce structural validity and feasibility, offering limited semantic-quality feedback for iterative refinement; incorporating evaluator-in-the-loop signals could better guide the refinement of activity chains and temporal schedules. We hope these directions will further strengthen workflow-based LLM agents for semantically coherent and physically plausible human mobility trajectory generation at scale.

\section*{Acknowledgments}
Research supported by the Intelligence Advanced Research
Projects Activity (IARPA) via the Department
of Interior/Interior Business Center (DOI/IBC) contract
number 140D0423C0033. The U.S. Government is authorized
to reproduce and distribute reprints for Governmental
purposes, notwithstanding any copyright annotation
thereon. Disclaimer: The views and conclusions contained
herein are those of the authors and should not
be interpreted as necessarily representing the official
policies or endorsements, either expressed or implied, of
IARPA or the U.S. Government.

\newpage
\bibliographystyle{IEEEtran}

\balance
\bibliography{references}

\end{document}